\pdfoutput=1

\documentclass[11pt]{article}

\usepackage{EMNLP2023}

\usepackage{times}
\usepackage{latexsym}

\usepackage[T1]{fontenc}

\usepackage[utf8]{inputenc}

\usepackage{microtype}
\usepackage{inconsolata}
\usepackage{graphicx}
\usepackage{longtable}
\usepackage{adjustbox}
\usepackage{tabularray}
\usepackage[utf8]{inputenc}
\usepackage[english,ukrainian]{babel}
\usepackage{tempora}

\title{The Grammar and Syntax Based Corpus Analysis Tool For The Ukrainian Language}

\author{Daria Stetsenko \\
  NASK National Research Institute \\ Warsaw, Poland \\
  \texttt{daria.stetsenko@nask.pl} \\\And
  Inez Okulska, Ph.D. \\
  NASK National Research Institute \\ Warsaw, Poland \\
  \texttt{inez.okulska@nask.pl} \\}

\begin{document}
\maketitle
\begin{abstract}
This paper provides an overview of a text mining tool the StyloMetrix developed initially for the Polish language and further extended for English and recently for Ukrainian. The StyloMetrix is built upon various metrics crafted manually by computational linguists and researchers from literary studies to analyze grammatical, stylistic, and syntactic patterns. 
The idea of constructing the statistical evaluation of syntactic and grammar features is straightforward and familiar for the languages like English, Spanish, German, and others; it is yet to be developed for low-resource languages like Ukrainian. We describe the StyloMetrix pipeline and provide some experiments with this tool for the text classification task. We also describe our package's main limitations and the metrics' evaluation procedure. 

\end{abstract}
\section{Introduction}

Ukrainian remains one of the low-resource languages with few practical applications in machine learning and deep learning. Many studies on the Ukrainian language are conducted in terms of multilingual settings, such as training the multilingual large language models \cite{laba2023contextual, mehta2023llm}, transformers \cite{panchenko2022ukrainian, choenni2020does}, or abstractive summarization \cite{galeshchuk2023abstractive}. We offer a corpus analysis tool for the Ukrainian language – the StyloMetrix\footnote{\url{https://github.com/ZILiAT-NASK/StyloMetrix}}. The underlying idea is not new in the NLP community but is recent in the Ukrainian language. 

This paper provides an overview of an open-source Python package – the StyloMetrix developed initially for the Polish language and further extended for English and recently for Ukrainian. The StyloMetrix is built upon a range of metrics crafted manually by computational linguists and researchers from literary studies to analyze stylometric features of texts from different genres. The principal purport of this package is to provide high-quality statistical evaluations of the general grammatical, lexical, and syntactic features of the text, regardless of its length, genre, or author. 

We organize our paper in the following way:
\begin{itemize}
    \item we provide an overview of similar tools for text analysis and a general idea of the corpus linguistics based on the syntactic and grammar representations;
    \item give an exhaustive characteristic of existing metrics for the Ukrainian language, their evaluation, and limitations;
    \item describe a case study with the StyloMetrix as the baseline model for the text classification task, providing the metrics analysis and feature importance of the classification model.
\end{itemize}

\section{Related Studies}

The idea to measure specific textual features to determine a text’s register or an author is not new. In 1998, \citeauthor{biber1998corpus} have developed a comprehensive methodological approach for corpus analysis based only on grammatical characteristics. \citet{biber2005corpus} argues that, although, semantic evaluations and descriptive analysis can provide a valuable insight about the narrative, it is not enough if one needs to discerne the genre of the text or to assess whether it belongs to a particular author and an epoch \cite{biber2005corpus}. On the other hand, grammatical/syntactic characteristics and figures of speech may come in handy and be less decisive and more exhaustive when it comes to genre, author or style estimation. M.A.K. Halliday supports this view and emphasises the general importance of corpus studies as a source of insight into the nature of language. He points out that "a language is inherently probabilistic and we need to extract the frequencies in the texts to establish probabilities in the grammatical system – not for the purpose of tagging and parsing, but to discover the interactions between different subsystems" \cite{aijmer2014english}.
 
The development of corpus-based grammar and syntactic tools for text mining has started in 1990s and is still an ongoing field of investigation. Some of the corpus-based techniques aim to manually study the English grammar and discourse. For instance, \citet{aarts1995livres} and \citet{mair1991quantitative} provide introductions on how to identify and extract syntactic and grammatical constructions in corpora to build tagging and parsing algorithms. They cover various aspects, limitations and boundaries related to grammar and syntax. Other researchers concentrate on specific incarnations of the language use. For example, \citet{tottie1991lexical} on negation and lexical diffusion in syntactic change; \citet{collins2006clefts} on prosody and pragmatics based on it-clefts and wh-clefts; \citet{granger1997automated} on automated retrieval of passives; \citet{mair1991quantitative} on infinitival complement clauses; and \citet{chomsky1988generative} has conducted the most valuable study on generative grammar that has served as a scaffold for contemporary natural language processing. Those techniques are the basis of modern tools and web-based services for text analysis. 

We follow the assumption that grammar and syntax can be enough for the tasks connected with style and author classification which are unified under the term stylometry \cite{10.1145/3132039}. 

The most popular applications for the stylometric analysis are the "Stylometry with R" (stylo) ~\cite{eder2016stylometry} (for English and Polish), WebSty~\cite{piaseck2018} and CohMetrix~\cite{graesser2004coh}. The stylo is a flexible R package for the high-level analysis of writing style in stylometry. The package can be applied at the supervised learning for the text classification \cite{eder2016stylometry}. WebSty~\cite{piaseck2018} is an accessible open-sourced library that encompasses  grammatical, lexical, and thematic parameters which can be manually selected by the user. The tool covers the Polish, English, German and Hungarian languages.  Coh-Metrix is a web-based platform that offers a wider range of descriptive statistic measurements. For example, low-level metrics counting pronouns per sentence, Text Easability Principal Component Scores, Referential Cohesion, LSA, Lexical Diversity, Connectives, Situation Model, Syntactic Complexity, Syntactic Pattern Density, Word Information, Readability, etc. \cite{mcnamara2014automated}. The documented versions of Coh-Metrix exist for Spanish \cite{quispesaravia2016coh}, Portuguese \cite{scarton2010coh}, and Chinese \cite{ouyang2021coh} (however, they are developed independently and not supported by the initial authors).

There are many tools for corpus analysis that look at concordances, n-grams, co-locations, key words and numerous frequency analysis which can be applied for the stylometric classification tasks, but most of them are quite primitive and basic with respect to the intricacy of grammar structures like tenses or syntactic phrases (the comprehensive list of tools can be accessed via the link in the footnote\footnote{\url{https://corpus-analysis.com/}}). 

Therefore, inspired by the powerful image of grammatical patterns and syntactic clauses we build the first (to our knowledge) corpus-analysis tool for the Ukrainian language that presents a thorough statistical evaluation of the Ukrainian grammar, syntactic patterns, and some descriptive lexical assessment. 

\begin{figure*}[!h]
\includegraphics[width=\textwidth]{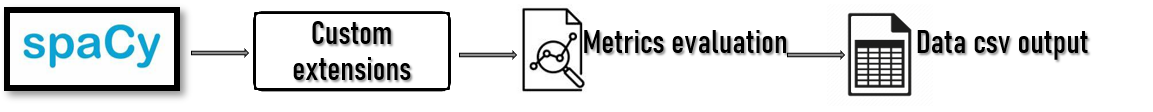}
\renewcommand{\figurename}{Figure}
\caption[english]{The pipeline of the StyloMetrix.}
\centering
\label{figure:SM}
\end{figure*}

\section{Grammatical Vector Representations}

\subsection{General outline}

The general pipeline of the tool is presented in the Figure. First, we utilize the standard spaCy pipeline of the transformer model for the Ukrainian language. The primary purpose of our package is not to build new tagger or parser algorithms but to add a higher level of grammatical and syntactic language characteristics and provide descriptive statistical measurements for each of them. Ukrainian is a fusional language, and the basic spaCy pipeline can trace only primary morphological features such as animacy, gender, case, number, aspect, degree, name type, verb form, and others\footnote{https://spacy.io/models/uk}. Nonetheless, these attributes do not cover all aspects of Ukrainian morphology and grammar, such as two types of conjugation, four types of declension, and present, past, and future tenses. Therefore, we leverage the last spaCy component of the pipeline and create custom extensions for each case\footnote{More about the metrics and their validation step is in subsection 3.3.}. Further, the tokens that fall under specific rules are calculated by the discussed formula at the stage of the Metric evaluation and are stored in the data frame that is further available for a user in the .csv format. As for the input – the StyloMetrix can be applied to any text length starting from a single sentence.

The StyloMetrix is a tool designed to calculate the mean value of a distinct grammar rule, a lexical component or a stylistic phenomenon. The statistical measurement is derived by the standard formula: $$\frac{\sum_{0}^{^{n}} w}{N}$$ where ${\sum_{0}^{^{n}} w}$  is the sum of all tokens that fall under the particular rule, and \textit{N} – is the total amount of tokens in the text. This evaluation holds for all metrics. Hence the output is acquired as a matrix, where the text instances are at the y-axis and the x-axis is the vectors of real numbers that stand for the specific metric. The obtained matrices can be utilized for various machine-learning tasks.

Primary developed for the Polish language, which is also fusional, the package has been used for stylometric analysis and text classification. For example, \citeauthor{okulskastyles} present a study on erotic vs. neutral text classification using the StyloMetrix vectors as the input to the RandomForest Classifier. The general accuracy has yielded around 0.90 score, giving us an impetus to deliver the primary metrics for the Ukrainian language and test their performance on the existing annotated datasets.  

\subsection{SpaCy Limitations}

\setlength{\tabcolsep}{10pt} 
\renewcommand{\arraystretch}{1.5} 
\begin{table*}[!h]
\centering
\scalebox{0.7}{%
\begin{tabular}{|l|l|l|l|}
\hline
\multicolumn{1}{|c|}{\textbf{Word}} &
  \multicolumn{1}{c|}{\textbf{spaCy tag}} &
  \multicolumn{1}{c|}{\textbf{Correct tag}} &
  \multicolumn{1}{c|}{\textbf{Sentence}} \\ \hline 
Закрапало &
  Aspect=Imp &
  Aspect=Perf &
  Із стріх закрапало, а з гір струмочки покотилися. \\[5pt] \hline
Веснянки &
  ADV &
  NOUN, Plural &
  Вже веснянки заспівали. \\ \hline
Замазалося &
  Aspect=Imp &
  Aspect=Perf &
  Високе небо замазалося зеленобурими хмарами, припало до землі, наче нагнітило на   неї. \\ \hline
Крук &
  Animacy=Inan &
  Animacy=Animate &
  Тільки чорний крук надувся, жалібно закрякав з високої могили серед пустельного   поля. \\ \hline
Завдання &
  Case=Acc &
  Case-Nom &
  Завдання буде зроблено. \\ \hline
Листа &
  Animacy=Anim &
  Animacy=Inan &
  Я напушу листа. \\ \hline
Осінню &
  \begin{tabular}[c]{@{}l@{}}ADJ       \\ Case=Acc\end{tabular} &
  \begin{tabular}[c]{@{}l@{}}NOUN\\ Case=Ins\end{tabular} &
  Повіває молодою осінню холодна річка з низів. \\ \hline
Низів &
  Case=Gen &
  Case=Loc &
  Повіває молодою осінню холодна річка з низів. \\ \hline
\end{tabular}%
}
\renewcommand{\tablename}{Table}
\caption{spaCy tags incongruencies.}
\label{tab:spacy}
\end{table*}

Before developing the rules for custom extensions and metrics, we verified the spaCy tags' correctness. Table~\ref{tab:spacy} presents the incongruencies which have been discerned. Among the most frequent misassignments are morphological features such as case, animacy, aspect, and gender. For example, "Ілля" is a typical male Ukrainian name that is tagged as feminine by the spaCy parser. Other inconsistencies are found in the part-of-speech annotation.

We intentionally highlight this part as it directly influences the quality of our metrics. Due to the probability of tags’ incorrectness, some tokens can be missing from the set; therefore, the final evaluation of the tool may be less precise. At the lexical level, we try to avoid this drawback by checking some explicit morphological characteristics through affixes or the position of a token in the sentence based on a dependency tree. The dependency tags have proven to be the most precise and robust. Hence we tend to rely on them more while implementing grammar and syntactic rules.

\subsection{Metrics assessment}

\begin{table}[!h]
\centering
\scalebox{0.9}{%
\begin{tabular}{|l|l|}
\hline
\multicolumn{1}{|c|}{\textbf{Group}} & \multicolumn{1}{c|}{\textbf{Amount}} \\ \hline
Lexical                              & 56 metrics                           \\ \hline
Grammar                              & 23 metrics                           \\ \hline
Syntax                               & 14 metrics                           \\ \hline
Part-of-speech                       & 12 metrics                           \\ \hline
\end{tabular}%
}
\renewcommand{\tablename}{Table}
\caption{Amount of metrics per group}
\label{tab:amm}

\end{table}
The Ukrainian version of the StyloMetrix incorporates 104 metrics subdivided into lexical forms, parts-of-speech incidence, and syntactic and grammatical structures. The complete list of metrics can be found in Appendix A. In this subsection, we strive to provide general descriptive characteristics and validation criteria for each group. 

Table~\ref{tab:amm} describes the number of metrics per category. With the StyloMetrix, academics can extract both conventional statistics of the text and features intrinsic to the Ukrainian language. For instance, the universal metrics are the type-token ratio, functional and content words, punctuation, and parts-of-speech statistics. A few examples are generated from the news text provided in Appendix B and presented below. 

\begin{itemize}
\item \textbf{L\_ADV\_POS}: [потрібно, відверто] – positive adverbs [needed, sincerely]
\item \textbf{L\_ANIM\_NOUN}: [Президент, агресор, людей] – animated nouns [President, aggressor, people]
\item \textbf{L\_DIRECT\_OBJ}: [час, нам, армію, потенціал, альтернативи, режим] – direct object [time, us, army, potential, alternatives, regime]
\item \textbf{L\_INDIRECT\_OBJ}: [світом, року, конференції, Україні, режимом] – indirect object (in Ukrainian denoted by case; in English translation we add prepositions) [(by) world, (during) a year, (at) a conference, (to) Ukraine, (in) regime]
\end{itemize}

Albeit the commonness of these measurements, it has been demonstrated by many researchers, e.g., the Coh-Metrix study, that these scores may provide valuable insight into the idiosyncratic characteristics of a text. 

To forms prominent in the Ukrainian language belong syntactic constructions such as parataxis, ellipses, and positioning (прикладка). Grammatical forms such as two types of the future tense, passive and active participles (дієприсливний доконаного \ недоконаного виду), adverbial perfect \ imperfect participles (дієприкметник доконаного \ недоконаного виду), four types of declensions, and seven cases. For instance:  

\begin{itemize}
    \item \textbf{SY\_PARATAXIS}: [Я, хотів, чути, від, світу, ", Україна, ,, ми, будемо, з, тобою, "]. – parataxis [I wanted to hear for the world: "Ukraine, we will stand with you".]
    \item \textbf{VF\_FIRST\_CONJ}: [затримка, підтримкою, помилкою, країна] – first declension [delay, support, mistake, country]
    \item \textbf{L\_GEN\_CASE}: [виступу, безпеки, лютого, життів, домовленостей] – genitive case (in Ukrainian denoted by suffix) [performance, safety, February, lives, agreements]
\end{itemize}

The examples are the raw outputs from the metrics, with added translation into English. We evaluate metrics based on the accuracy score assessed by the trained linguist. The best accuracy has been achieved in the part-of-speech metrics – 0.957, due to their reliance on the spaCy tagger. The lexical metrics have obtained a weighted accuracy of - 0.934. Some discrepancies have occurred at relative and superlative adjectives, adverbs, and case misalignment because of the tagger performance. The grammar group scored 0.912; the inconsistency has occurred in declensions metrics. The syntactic group has got 0.886 in light of the complex constructions, such as parataxis and positioning, that may produce incongruencies.  

The accuracy scores indicate that the metrics perform well overall but have some limitations in dealing with complex structures. As the StyloMetrix provides each metric's mean value, a researcher can skip looking into Ukrainian texts to extract the necessary features and conduct further analysis based on the obtained statistics. The descriptions are available for every metric, some with external links to the Universal Dependencies project\footnote{\url{https://universaldependencies.org/}}.  

\section{Experiments}

\begin{table*}[!h]
\centering
\scalebox{1.0}{%
\begin{tabular}{|l|ll|}
\hline
\multicolumn{1}{|c|}{\textbf{Model}} & \multicolumn{2}{c|}{\textbf{Large training set}} \\ \hline
                     & \multicolumn{1}{l|}{\textbf{Paper}} & \textbf{Our Result} \\ \hline
NB-SVM               & \multicolumn{1}{l|}{0.64}           & -                   \\ \hline
\textbf{SM-Voting Classifier} & \multicolumn{1}{l|}{-}              & \textbf{0.66}                \\ \hline
Ukr-RoBERTa          & \multicolumn{1}{l|}{0.75}           & 0.82                \\ \hline
Ukr-ELECTRA          & \multicolumn{1}{l|}{0.72}           & 0.89                \\ \hline
\end{tabular}%
}
\renewcommand{\tablename}{Table}
\caption{Results of our experiments compared to the paper by Panchenko et al.}
\label{tab:results}
\end{table*}

The primary endeavor behind the StyloMetrix project is to develop a tool that can produce plausible descriptive results for any text, regardless of its length or genre. Typically, researchers need to provide the baseline when they want to evaluate new or re-trained models. So to say the “dummy” algorithm, which serves as the lower bound, must be surpassed by another model to prove its efficiency\cite{1011452738037}. However, should the baseline model be “dummy”? This section attempts to represent the StyloMetrix as a “canny” baseline for text classification tasks. We further illustrate how to analyze the StyloMetrix baseline model and the possibility of making beneficial inferences about the data based solely on its output.

Conducting a supervised text classification in the Ukrainian language remains challenging due to the scarcity of labeled datasets. There exist a few open-source corpora which can be relevant to this task. For instance, the largest and most popular corpus known by now is UberText 2.0 \cite{chaplynskyi-2023-introducing}. The data is subdivided into five smaller datasets: the news dataset, which incorporates short news, longer articles, interviews, opinions, and blogs scraped from 38 news websites; the fiction dataset, with novels, prose, and poetry; the social dataset, covers 264 public telegram channels; the Wikipedia corpus; and the court dataset with decisions of the Supreme Court of Ukraine. The UA news corpus\footnote{\url{https://github.com/fido-ai/ua datasets/blob/main/ua_datasets/src/text_classification/README.md}} is a collection of over 150 thousand news articles from more than 20 news resources. Dataset samples are divided into five categories: politics, sport, news, business, and technologies. UA-SQuAD is a Ukrainian version of Stanford Question Answering Dataset\footnote{\url{https://github.com/fido-ai/ua-datasets/blob/main/ua_datasets/src/question_answering/README.md}}, and UA-GEC: Grammatical Error Correction and Fluency Corpus for the Ukrainian language \cite{syvokon2021uagec}. The list with all state-of-the-art datasets can be found via the link in the footnote\footnote{\url{https://github.com/asivokon/awesome-ukrainian-nlp/blob/master/README.md}}.

We ground our experiments on the well-established benchmark public dataset\footnote{\url{https://www.kaggle.com/competitions/ukrainian-news-classification/data}} provided by Kaggle project. The corpus has been scrapped from seven Ukrainian news websites: BBC News Ukraine, NV (New Voice Ukraine), Ukrainian Pravda, Economic Pravda, European Pravda, Life Pravda, and Unian. Ukrainian computer scientists \citeauthor{panchenko2022ukrainian} have developed the described corpus. The researchers give an exhaustive outlook on the data preprocessing steps and the number of texts in the train/test split (57789/ 24765, respectively). The Kaggle platform offers two training splits from the existing sample: large (57460) and small (9299). In their paper, the academics demonstrate their models' performance scores on the two training splits discussed\cite{panchenko2022ukrainian}. We are left with the training splits because we cannot use the initial train and test split as it is unavailable to the public. 

The large training data partially incorporates the small training sample; hence we leverage the larger corpus, subdividing it into 80/20\% training and testing samples, with 15\% for validation. The obtained results are evaluated with macro-averaged F1-score, the same criterion as in the paper. The baseline model leveraged in the study by \citeauthor{panchenko2022ukrainian} was Naïve Bayes with SVM; we have added the StyloMetrix with Voting Classifier as our baseline. The Voting Classifier is composed of RandomForest, AdaBoost, and Logistic Regression.  As for the main models, we keep the ones utilized in the paper: ukr-RoBERTa\cite{minixhofer-etal-2022-wechsel} and ukr-ELECTRA\cite{stefan_schweter_2020_4267880}.

As inferred from Table~\ref{tab:results}, the StyloMetrix-Voting Classifier has scored higher than the Naïve Bayes – SVM model but not much, which allows it to serve as a baseline for other more advanced algorithms.

\begin{figure*}[!h]
\includegraphics[width=\textwidth]{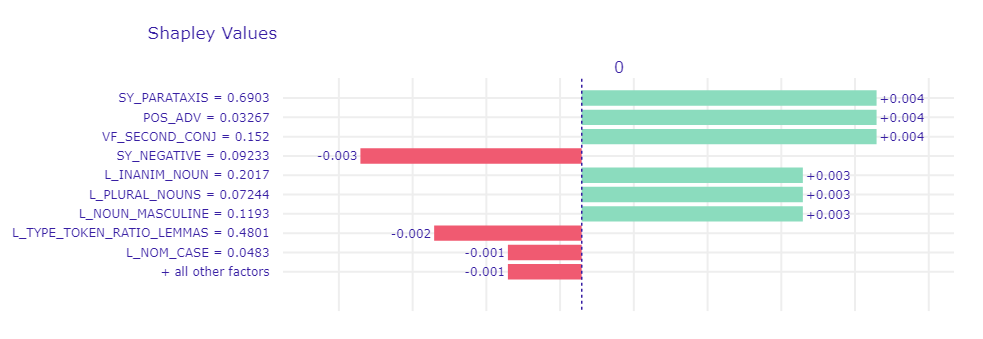}
\renewcommand{\figurename}{Figure}
\caption[english]{The Shapley values for class 0.}
\centering
\label{figure:shapley}
\end{figure*}

\subsection{Model Explanation}

Unlike the Naïve Bayes – SVM model, the StyloMetrix offers the possibility to extract the descriptive statistics for each group, looking at the most discriminative metrics. For example, we have chosen class 0 -  BBC News Ukraine, to describe the possible data analysis approaches with the StyloMetrix.

One of the wide-used methods of explainable AI is Shapley values\cite{hart1989shapley}, which shows the average marginal contributions of features. To make the most of this type of explanation, Shapley values are usually applied to features that can be reversely interpreted, such as categorical values, or to anomaly detection~\cite{tallon2020explainable}. The most common text representation offering static or dynamic embeddings like GloVe, Word2Vec, or BERT-based vectors does not allow human interpretation of such explanations. The Shapley values indicate the most essential features locally and globally, but the features themselves remain some random columns. With StyloMetrix, on the other hand, the text vector representation consists of interpretable values: each element of the vector translates directly into a given linguistic metric. In this case, indicating the local or global contribution of top features allows for linguistic analysis of grammar or lexical patterns that impact the model's decision when predicting the class. 

To implement this, we leverage an open-source library – DALEX \cite{JMLR:v22:20-1473}. As shown in Figure~\ref{figure:shapley} the metrics’ significance for class 0 based on their contributions to the model's decision is described. Ultimately, the syntactic metric for parataxis, adverbs, second declension, inanimate nouns, plural nouns, and masculine nouns are prominent in texts that belong to class 0. Vice versa, negative sentences, type-token ratio, and nominative case lower the likelihood of a text falling under this category.

We can dive even deeper into the text statistics and extract the aggregated mean values of the metrics from the StyloMetrix output. As the final vectors are saved in the .csv file, it is easy to find the needed metric and estimate the average mean value for the class. For instance, based on the obtained Shapley, we provide the metrics description and average mean of all texts under class 0 (Table~\ref{tab:mean}). This, in turn, serves the linguistic analysis offering a statistical baseline for a given text genre, including a wide range of metrics. Especially in a multi-class classification, it is vital to compare the baseline against other genres (classes) and draw conclusions about local and global distinctive features.

\begin{table*}[]
\centering
\scalebox{0.8}{%
\begin{tabular}{|c|l|l|}
\hline
\textbf{Metric} & \multicolumn{1}{c|}{\textbf{Description}} & \multicolumn{1}{c|}{\textbf{Mean}} \\ \hline
SY\_PARATAXIS                 & Number of words in sentences with parataxis  & 0,02731253208958335    \\ \hline
POS\_ADV                      & Incidence of adverbs                         & 0,04840493864411134    \\ \hline
VF\_SECOND\_CONJ              & Incidence of words in the second declension  & 0,00027041364427593664 \\ \hline
SY\_NEGATIVE                  & Incidence of words in the negative sentences & 0,08136554265171815    \\ \hline
L\_INANIM\_NOUNS              & Incidence of inanimate nouns                 & 0,013161611405387586   \\ \hline
L\_PLURAL\_NOUNS              & Incidence of plural nouns                    & 0,002385407919553026   \\ \hline
L\_NOUN\_MASCULINE            & Incidence of masculine nouns                 & 0,20552832194690895    \\ \hline
L\_TYPE\_TOKEN\_RATIO\_LEMMAS & Type-token ratio for words lemmas            & 0,054029343395248786   \\ \hline
L\_NOM\_CASE                  & Incidence of nouns in Nominative case        & 0,06036709344834804    \\ \hline
\end{tabular}%
}
\renewcommand{\tablename}{Table}
\caption{The StyloMetrix mean values and descriptions of each metric. }
\label{tab:mean}
\end{table*}

Therefore, we can conclude that the StyloMetrix as the baseline “cunny” model can bring some beneficial insights about the texts and the significance of the metrics for a particular classification model. We have presented only one approach to data evaluation with the XAI tool. There are other possibilities to research this area and expand the horizon of the StyloMetrix application and existing metrics.  

\section{Conclusions}
Albeit the idea of constructing the statistical measures of syntactic and grammar features of the text is not new, the experiments discussed in this paper highlight the relevance and significance of creating open-source packages like the StyloMetrix. In the article, we have outlined the main metrics available in the tool's current version and provided some descriptive analysis with the StyloMetrix. We have also discussed the applicability of the corpus analysis tool like StyloMetrix as the baseline "cunny" model for machine learning.

Through experiments, we have traced the metrics importance in a model for classification tasks using the XAI tool – DALEX. More rigorous and detailed analysis is yet to be done in this field, and we consider it the next milestone for our study. The metrics have performed well at the validation step and can be efficient for linguistic analysis of different text genres and the cross-linguistic analysis with other languages such as Polish and English (also available in the StyloMetrix).

\section*{Limitations}
The main stumbling pint to obtain the StyloMetrix representations is RAM and GPU access. Although the tool does not need high RAM it is still preferably to utilize one; and as the StyloMetrix is grounded on the spaCy model, it is preferably to install both the English and the Ukrainian transformer models, which require the GPU. 

Moreover the metrics may output some inconsistencies as it has been described in the main part of this paper. 

\section*{Ethics Statement}
The work presented in this paper aims to advance the field of natural language processing by developing novel methods and applications. However, we acknowledge that our work may also have broader social and ethical implications that need to be considered and addressed.
Some of the potential positive impacts of our work are:
\begin{itemize}

\item It can improve the quality and accessibility of information and communication for various domains and languages, especially for low-resource languages.
\item It can enhance the efficiency of neural network models for text classification, such as search engines, chatbots, translators, summarizers, etc.
\item It can foster new insights and discoveries in linguistics, cognitive science, artificial intelligence, and other related disciplines.
Some of the potential negative impacts of our work are:
\item It may introduce biases to the neural network models due to inconsistent parsing and tagging the data. 
To mitigate these risks and ensure that our work is conducted in a responsible and ethical manner, we adhere to the following principles and practices:
\item We follow the relevant ethical guidelines and codes of conduct.
\item We conduct thorough literature reviews and background checks to identify and acknowledge
the existing work and contributions in our field.
\item We collect, process, store, and share natural language data and models in accordance with
the best practices for data management and protection,
\item We evaluate and report the performance and limitations of our natural language processing tool in a transparent and rigorous way, using appropriate metrics and benchmarks, and disclosing any potential sources of error or uncertainty.
    \item  We consider and anticipate the possible impacts and implications of our natural language processing systems or services on different stakeholders, such as users, clients, partners, competitors, regulators, or society at large.

\end{itemize}

\bibliography{emnlp}
\bibliographystyle{plainnat}

\appendix
\section{Example Appendix}
\label{sec:appendix}

\begin{table}[!h]
\centering
\scalebox{0.8}{%
\begin{tabular}{|l|l|}
\hline
\multicolumn{1}{|c|}{\textbf{Metric}} & \multicolumn{1}{c|}{\textbf{Description}} \\ \hline
POS\_VERB                             & Incidence of Verbs                        \\ \hline
POS\_NOUN                             & Incidence of Nouns                        \\ \hline
POS\_ADJ                              & Incidence of Adjectives                   \\ \hline
POS\_ADV                              & Incidence of Adverbs                      \\ \hline
POS\_DET                              & Incidence of Determiners                  \\ \hline
POS\_INTJ                             & Incidence of Interjections                \\ \hline
POS\_CONJ                             & Incidence of Conjunctions                 \\ \hline
POS\_PART                             & Incidence of Particles                    \\ \hline
POS\_NUM                              & Incidence of Numerals                     \\ \hline
POS\_PREP                             & Incidence of Prepositions                 \\ \hline
POS\_PRO                              & Incidence of Pronouns                     \\ \hline
POS\_OTHER                            & Incidence of Code-Switching               \\ \hline
\end{tabular}%
}
\renewcommand{\tablename}{Table}
\caption{Part-of-speech group }
\label{tab:pos}
\end{table}

\begin{table}[!h]
\centering
\scalebox{0.9}{%
\begin{tabular}{|l|l|}
\hline
\multicolumn{1}{|c|}{\textbf{Metric}} & \multicolumn{1}{c|}{\textbf{Description}} \\ \hline
L\_PRON\_RELATIVE                     & Incidence of relative pronoun 'що'        \\ \hline
L\_PRON\_RFL                          & Incidence of reflexive pronoun            \\ \hline
L\_PRON\_TOT                          & Incidence of total pronouns               \\ \hline
L\_QUALITATIVE\_ADJ\_SUP              & Incidence of qualitative superlative adj  \\ \hline
L\_QULITATIVE\_ADJ\_P                 & Incidence of qualitative adj positive     \\ \hline
L\_RELATIVE\_ADJ                      & Incidence of relative adj                 \\ \hline
L\_SURNAMES                           & Incidence of surnames                     \\ \hline
\end{tabular}%
}
\renewcommand{\tablename}{Table}
\caption{Lexical metrics}
\label{tab:lex3}
\end{table}

\begin{table}[!h]
\centering
\scalebox{0.9}{%
\begin{tabular}{|l|l|}
\hline
\multicolumn{1}{|c|}{\textbf{Metric}} & \multicolumn{1}{c|}{\textbf{Description}} \\ \hline
L\_PUNCT                              & Incidence of punctuation                  \\ \hline
L\_PUNCT\_DOT                         & Incidence of dots                         \\ \hline
L\_PUNCT\_COM                         & Incidence of comma                        \\ \hline
L\_PUNCT\_SEMC                        & Incidence of semicolon                    \\ \hline
L\_PUNCT\_COL                         & Incidence of colon                        \\ \hline
L\_PUNCT\_DASH                        & Incidence of dashes                       \\ \hline
\end{tabular}%
}
\renewcommand{\tablename}{Table}
\caption{Lexical metrics for punctuation.}
\renewcommand{\tablename}{Table}
\label{tab:punct}
\end{table}

\setlength{\tabcolsep}{10pt} 
\renewcommand{\arraystretch}{1.5} 
\begin{table*}[!h]
\centering
\scalebox{0.8}{%
\begin{tabular}{|l|l|}
\hline
\multicolumn{1}{|c|}{\textbf{Metric}} & \multicolumn{1}{c|}{\textbf{Description}}                                   \\ \hline
VF\_ROOT\_VERB\_IMPERFECT             & Root verbs and conjunctions in imperfect aspect                             \\ \hline
VF\_ALL\_VERB\_IMPERFECT              & Incidence of all verbs in imperfect aspect                                  \\ \hline
VF\_ROOT\_VERB\_PERFECT               & Root verbs and conjunctions in perfect aspect                               \\ \hline
VF\_ALL\_VERB\_PERFECT                & Incidence of all verbs in perfect aspect                                    \\ \hline
VF\_PRESENT\_IND\_IMPERFECT     & Incidence of verbs in the present tense, indicative mood, imperfect aspect                  \\ \hline
VF\_PAST\_IND\_IMPERFECT              & Incidence of verbs in the past tense, indicative mood, imperfect aspect     \\ \hline
VF\_PAST\_IND\_PERFECT                & Incidence of verbs in the past tense, indicative mood, perfect aspect       \\ \hline
VF\_FUT\_IND\_PERFECT                 & Incidence of verbs in the future tense, indicative mood, perfect aspect     \\ \hline
VF\_FUT\_IND\_IMPERFECT\_SIMPLE & Incidence of verbs in the future tense, indicative mood, imperfect aspect, simple verb form \\ \hline
VF\_FUT\_IND\_COMPLEX                 & Incidence of verbs in the future tense, indicative mood, complex verb forms \\ \hline
VT\_FIRST\_CONJ                       & Incidence of verbs in the first declension                                  \\ \hline
VT\_SECOND\_CONJ                      & Incidence of verbs in the second declension                                 \\ \hline
VT\_THIRD\_CONJ                       & Incidence of verbs in the third declension                                  \\ \hline
VT\_FOURTH\_CONJ                      & Incidence of verbs in the fourth declension                                 \\ \hline
VF\_TRANSITIVE                        & Incidence of transitive verbs                                               \\ \hline
VF\_PASSIVE                           & Incidence of verbs in the passive form                                      \\ \hline
VF\_PARTICIPLE\_PASSIVE               & Incidence of passive participles                                            \\ \hline
VF\_PARTICIPLE\_ACTIVE                & Incidence of active participles                                             \\ \hline
VF\_INTRANSITIVE                      & Incidence of intransitive verbs                                             \\ \hline
VF\_INFINITIVE                        & Incidence of verbs in infinitive                                            \\ \hline
VF\_IMPERSONAL\_VERBS                 & Incidence of impersonal verbs                                               \\ \hline
VF\_ADV\_PRF\_PART                    & Incidence of adverbial perfect participles                                  \\ \hline
VF\_ADV\_IMPRF\_PART                  & Incidence of adverbial imperfect participles                                \\ \hline
\end{tabular}%
}
\renewcommand{\tablename}{Table}
\caption{Grammar group}
\label{tab:grammar}
\end{table*}

\setlength{\tabcolsep}{20pt} 
\renewcommand{\arraystretch}{1.5} 
\begin{table}[!h]
\centering
\scalebox{1.0}{%
\begin{tabular}{|l|l|}
\hline
\multicolumn{1}{|c|}{\textbf{Metric}} & \multicolumn{1}{c|}{\textbf{Description}}         \\ \hline
SY\_PARATAXIS                         & Number of words in parataxis sentences            \\ \hline
SY\_DIRECT\_SPEECH                    & Number of words in direct speech                  \\ \hline
SY\_NEGATIVE                          & Number of words in negative sentences             \\ \hline
SY\_NON\_FINITE                       & Number of words in sentences without any verbs    \\ \hline
SY\_QUOTATIONS                        & Number of words in sentences with quotation marks \\ \hline
SY\_EXCLAMATION                       & Number of words in exclamatory sentences          \\ \hline
SY\_QUESTION                          & Number of words in interrogative sentences        \\ \hline
SY\_ELLIPSES                          & Number of words in elliptic sentences             \\ \hline
SY\_POSITIONING                       & Number of positionings (прикладка)                \\ \hline
SY\_CONDITIONAL                       & Number of words in conditional sentences          \\ \hline
SY\_IMPERATIVE                        & Number of words in imperative sentences           \\ \hline
SY\_AMPLIFIED\_SENT                   & Number of words in amplified sentences            \\ \hline
SY\_NOUN\_PHRASES                     & Number of noun phrases                            \\ \hline
\end{tabular}%
}
\renewcommand{\tablename}{Table}
\caption{Syntactic group}
\label{tab:synt}
\end{table}

\begin{table}[!h]
\centering
\scalebox{0.9}{%
\begin{tabular}{|l|l|}
\hline
\multicolumn{1}{|c|}{\textbf{Metric}} & \multicolumn{1}{c|}{\textbf{Description}} \\ \hline
L\_TYPE\_TOKEN\_RATIO\_LEMMAS         & Type-token ratio for words lemmas         \\ \hline
L\_CONT\_A                            & Incidence of Content words                \\ \hline
L\_FUNC\_A                            & Incidence of Function words               \\ \hline
L\_CONT\_T                            & Incidence of Content words types          \\ \hline
L\_FUNC\_T                            & Incidence of Function words types         \\ \hline
L\_PLURAL\_NOUNS                      & Incidence of nouns in plural              \\ \hline
L\_SINGULAR\_NOUNS                    & Incidence of nouns in singular            \\ \hline
L\_PROPER\_NAME                       & Incidence of proper names                 \\ \hline
L\_PERSONAL\_NAME                     & Incidence of personal names               \\ \hline
L\_NOM\_CASE                          & Incidence of nouns in Nominative case     \\ \hline
L\_GEN\_CASE                          & Incidence of nouns in Genitive case       \\ \hline
L\_DAT\_CASE                          & Incidence of nouns in Dative case         \\ \hline
L\_ACC\_CASE                          & Incidence of nouns in Accusative case     \\ \hline
L\_INS\_CASE                          & Incidence of nouns in Instrumental case   \\ \hline
L\_LOC\_CASE                          & Incidence of nouns in Locative case       \\ \hline
L\_VOC\_CASE                          & Incidence of nouns in Vocative case       \\ \hline
L\_INDIRECT\_ADJ                      & Incidence of indirect adjective           \\ \hline
\end{tabular}%
}
\renewcommand{\tablename}{Table}
\caption{Lexical metrics}
\label{tab:lex1}
\end{table}

\begin{table*}[!h]
\centering
\scalebox{1.0}{%
\begin{tabular}{|l|l|}
\hline
\multicolumn{1}{|c|}{\textbf{Metric}} & \multicolumn{1}{c|}{\textbf{Description}} \\ \hline
L\_DIRECT\_ADJ                        & Incidence of direct adjective             \\ \hline
L\_QUALITATIVE\_ADJ\_SUP              & Incidence of qualitative superlative adj  \\ \hline
L\_QUALITATIVE\_ADJ\_CMP              & Incidence of relative adj                 \\ \hline
L\_RELATIVE\_ADJ                      & Incidence of relative adj                 \\ \hline
L\_QULITATIVE\_ADJ\_P                 & Incidence of qualitative adj positive     \\ \hline
L\_ANIM\_NOUN                         & Incidence of animated nouns               \\ \hline
L\_ADV\_CMP                           & Incidence of comparative adverbs          \\ \hline
L\_ADV\_POS                           & Incidence of positive adverbs             \\ \hline
L\_ADV\_SUP                           & Incidence of superlative adverbs          \\ \hline
L\_DIMINUTIVES                        & Incidence of diminutives                  \\ \hline
L\_FEMININE\_NAMES                    & Incidence of feminine proper nouns        \\ \hline
L\_FLAT\_MULTIWORD                    & Incidence of flat multiwords expressions  \\ \hline
L\_INANIM\_NOUN                       & Incidence of inanimate nouns              \\ \hline
L\_GIVEN\_NAMES                       & Incidence of given names                  \\ \hline
L\_MASCULINE\_NAMES                   & Incidence of masculine proper nouns       \\ \hline
L\_NOUN\_MASCULINE                    & Incidence of masculine nouns              \\ \hline
L\_NOUN\_FAMININE                     & Incidence of feminine nouns               \\ \hline
L\_NOUN\_NEUTRAL                      & Incidence of neutral nouns                \\ \hline
L\_NUM\_CARD                          & Incidence of numerals cardinals           \\ \hline
L\_NUM\_ORD                           & Incidence of numerals ordinals            \\ \hline
L\_PRON\_DEM                          & Incidence of demonstrative pronouns       \\ \hline
L\_PRON\_INT                          & Incidence of indexical pronouns           \\ \hline
L\_PRON\_NEG                          & Incidence of negative pronoun             \\ \hline
L\_PRON\_POS                          & Incidence of possessive pronoun           \\ \hline
L\_PRON\_PRS                          & Incidence of personal pronouns            \\ \hline
L\_PRON\_REL                          & Incidence of relative pronouns            \\ \hline
\end{tabular}%
}
\renewcommand{\tablename}{Table}
\caption{Lexical metrics}
\label{tab:lex2}
\end{table*}

\end{document}